\documentclass[sigconf,screen]{acmart}

\AtBeginDocument{%
  \providecommand\BibTeX{{%
    \normalfont B\kern-0.5em{\scshape i\kern-0.25em b}\kern-0.8em\TeX}}}

\copyrightyear{2022} 
\acmYear{2022} 
\setcopyright{acmcopyright}
\acmConference[MM '22]{Proceedings of the 30th ACM International Conference on Multimedia}{October 10--14, 2022}{Lisboa, Portugal} 
\acmBooktitle{Proceedings of the 30th ACM International Conference on Multimedia (MM '22), October 10--14, 2022, Lisboa, Portugal} 
\acmPrice{15.00} 
\acmDOI{10.1145/3503161.3548164} \acmISBN{978-1-4503-9203-7/22/10}

\usepackage{subfigure}
\usepackage{multirow}
\usepackage{colortbl} 
\usepackage{dsfont}
\usepackage{arydshln}

\usepackage{xcolor}
\definecolor{ForestGreen}{RGB}{34,139,34}

\newcommand{\Mat}[1]{\mathbf{#1}}

\usepackage{pifont}

\newcommand{\ie}{\emph{i.e., }}
\newcommand{\eg}{\emph{e.g., }}

\newcommand{\cf}{\emph{cf. }}
\newcommand{\vs}{\emph{v.s. }}

\definecolor{mygray-bg}{gray}{0.9}

\newcommand{\tabincell}[2]{\begin{tabular}{@{}#1@{}}#2\end{tabular}}

\begin{document}

\title{Integrating Object-aware and Interaction-aware Knowledge for Weakly Supervised Scene Graph Generation}



\author{Xingchen Li}
\affiliation{%
  \institution{Zhejiang University \country{China}}
}
\email{xingchenl@zju.edu.cn}

\author{Long Chen}
\authornote{Corresponding author.}
\affiliation{%
  \institution{Columbia University \country{USA}}
}
\email{zjuchenlong@gmail.com}

\author{Wenbo Ma}
\affiliation{%
  \institution{Zhejiang University \country{China}}
}
\email{mwb@zju.edu.cn}

\author{Yi Yang}
\affiliation{%
 \institution{Zhejiang University \country{China}}
}
\email{yee.i.yang@gmail.com}

\author{Jun Xiao}
\affiliation{%
  \institution{Zhejiang University \country{China}}
}
\email{junx@cs.zju.edu.cn}

\renewcommand{\shortauthors}{Xingchen Li et al.}

\begin{abstract}
  Recently, increasing efforts have been focused on Weakly Supervised Scene Graph Generation (WSSGG). The mainstream solution for WSSGG typically follows the same pipeline: they first align text entities in the weak image-level supervisions (\eg unlocalized relation triplets or captions) with image regions, and then train SGG models in a fully-supervised manner with aligned instance-level ``pseudo'' labels. However, we argue that most existing WSSGG works only focus on \emph{object-consistency}, which means the grounded regions should have the same object category label as text entities. While they neglect another basic requirement for an ideal alignment: \emph{interaction-consistency}, which means the grounded region pairs should have the same interactions (\ie visual relations) as text entity pairs. Hence, in this paper, we propose to enhance a simple grounding module with both object-aware and interaction-aware knowledge to acquire more reliable pseudo labels. To better leverage these two types of knowledge, we regard them as two teachers and fuse their generated targets to guide the training process of our grounding module. Specifically, we design two different strategies to adaptively assign weights to different teachers by assessing their reliability on each training sample. Extensive experiments have demonstrated that our method consistently improves WSSGG performance on various kinds of weak supervision\footnote{Code will be available at \url{https://github.com/xcppy/WS-SGG}}. 

\end{abstract}




\maketitle

\begin{figure}
    \centering
    \includegraphics[width=0.95\linewidth]{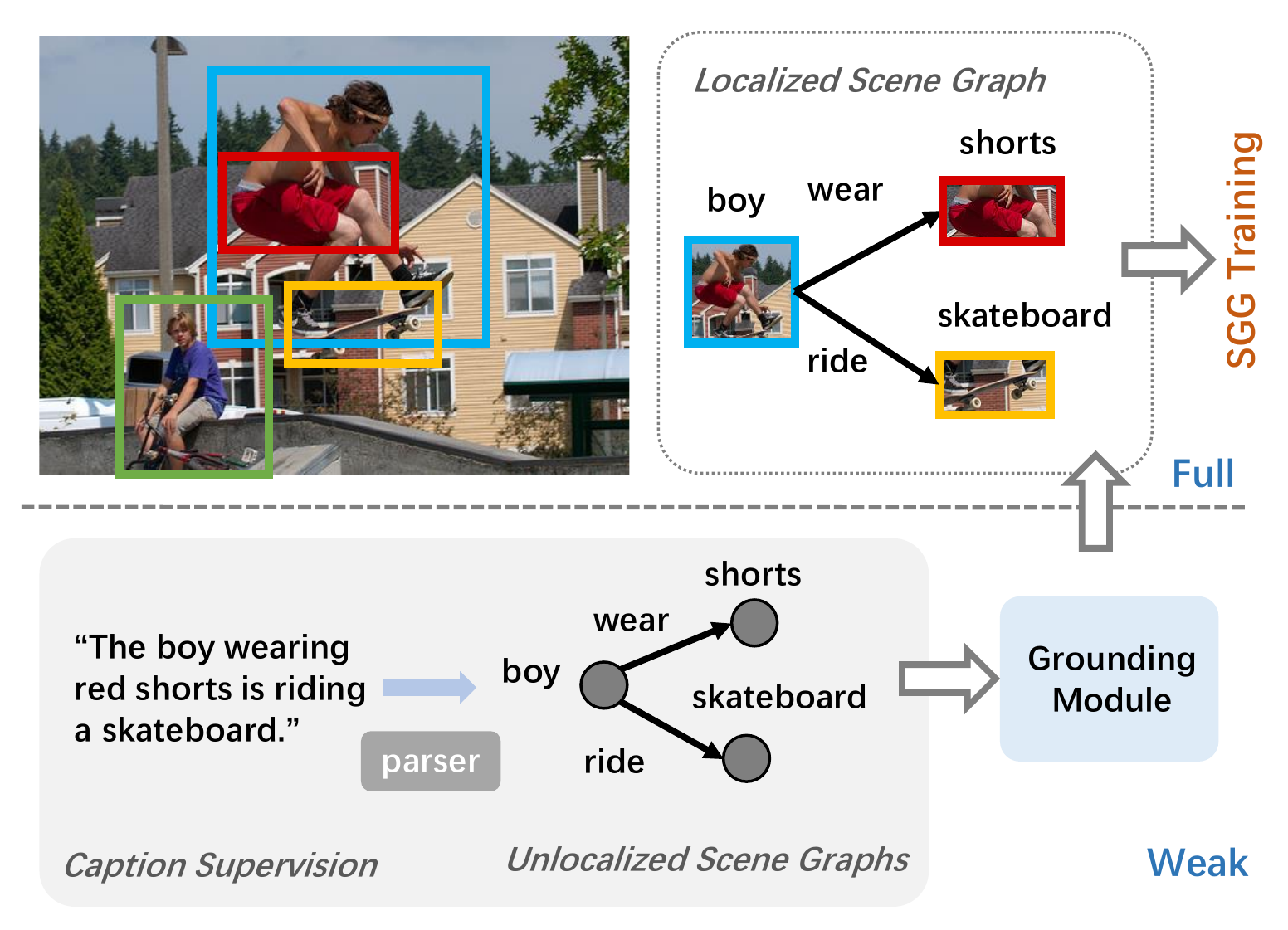}
    \caption{\emph{Top}: Full supervision in the form of localized scene graph for SGG. \emph{Bottom}: Weak supervision forms for WSSGG. WSSGG can be transformed into fully-supervised SGG by a grounding module to align text entities to image regions.}
    \label{fig:intro}
    \Description[Introduction of SGG Supervision.]{Introduction of SGG Supervision.}
\end{figure}

\section{Introduction}

Scene graph is a visually-grounded graph structured representation for an image scene: the nodes represent objects in an image and the edges represent pairwise visual relationships (\ie predicates) between two objects.
Due to its inherent interpretability, scene graph is able to support many downstream vision-and-language tasks, such as image retrieval~\cite{johnson2015image, wang2020cross, schroeder2020structured}, image captioning~\cite{yang2020hierarchical, li2019know, yang2019auto, chen2017sca, chen2021human, liu2021region, wang2021high, xu2019multi}, and visual grounding~\cite{jing2020visual,chen2021ref}.
Unfortunately, fully-supervised Scene Graph Generation (SGG) approaches require tons of instance-level annotations, in the form of localized visual relation triplets (\ie \texttt{<subject, predicate, object>}) with bounding box annotations for both \texttt{subject} and \texttt{object} (\cf Figure~\ref{fig:intro} top).
Such instance-level annotations for large-scale scene graph datasets require excessive human labeling efforts.
Accordingly, recent researches begin to focus on Weakly Supervised Scene Graph Generation (\textbf{WSSGG}) and looking for more accessible and cheap supervision for both objects and relationships.

Existing WSSGG works~\cite{zhang2017ppr,zareian2020weakly,ye2021linguistic, shi2021simple} mainly consider two types of image-level weak supervision: 1)~\emph{unlocalized scene graphs}~\cite{ye2021linguistic, shi2021simple}: a set of relationship triplets without corresponding bounding boxes annotations (\cf Figure~\ref{fig:intro}). 2)~\emph{Image captions}: sentences describe the objects and their interactions existing in an image scene (\cf Figure~\ref{fig:intro}).
Compared to unlocalized scene graphs, captions are easier to be collected from the Web for a large-scale dataset construction.
Current caption-based WSSGG works~\cite{ye2021linguistic,zhong2021learning} usually exploit a pretrained language parser~\cite{schuster2015generating} to convert each caption into a set of unlocalized relation triplets (\cf Figure~\ref{fig:intro}).
Although both two kinds of image-level weak supervisions mentioned above can greatly reduce the annotation cost, there is no free lunch, they inevitably increase the difficulty of SGG model training.

Currently, the mainstream solution for WSSGG~\cite{ye2021linguistic,shi2021simple,zareian2020weakly} typically trains a grounding module to assign ``pseudo'' instance-level labels for each image region pair and transforms WSSGG into fully-supervised SGG, as shown in Figure~\ref{fig:intro}.
Under such a pipeline, the challenge is to find the correct alignment between the text entities in unlocalized relation triplets (\ie \texttt{subject} and \texttt{object}) and image regions.
To accomplish this, most existing WSSGG works attempt to train the grounding module under the weak supervision of aligned image-triplets pairs~\cite{zareian2020weakly, shi2021simple, ye2021linguistic}.

\begin{figure}[t]
    \centering
    \includegraphics[width=0.95\linewidth]{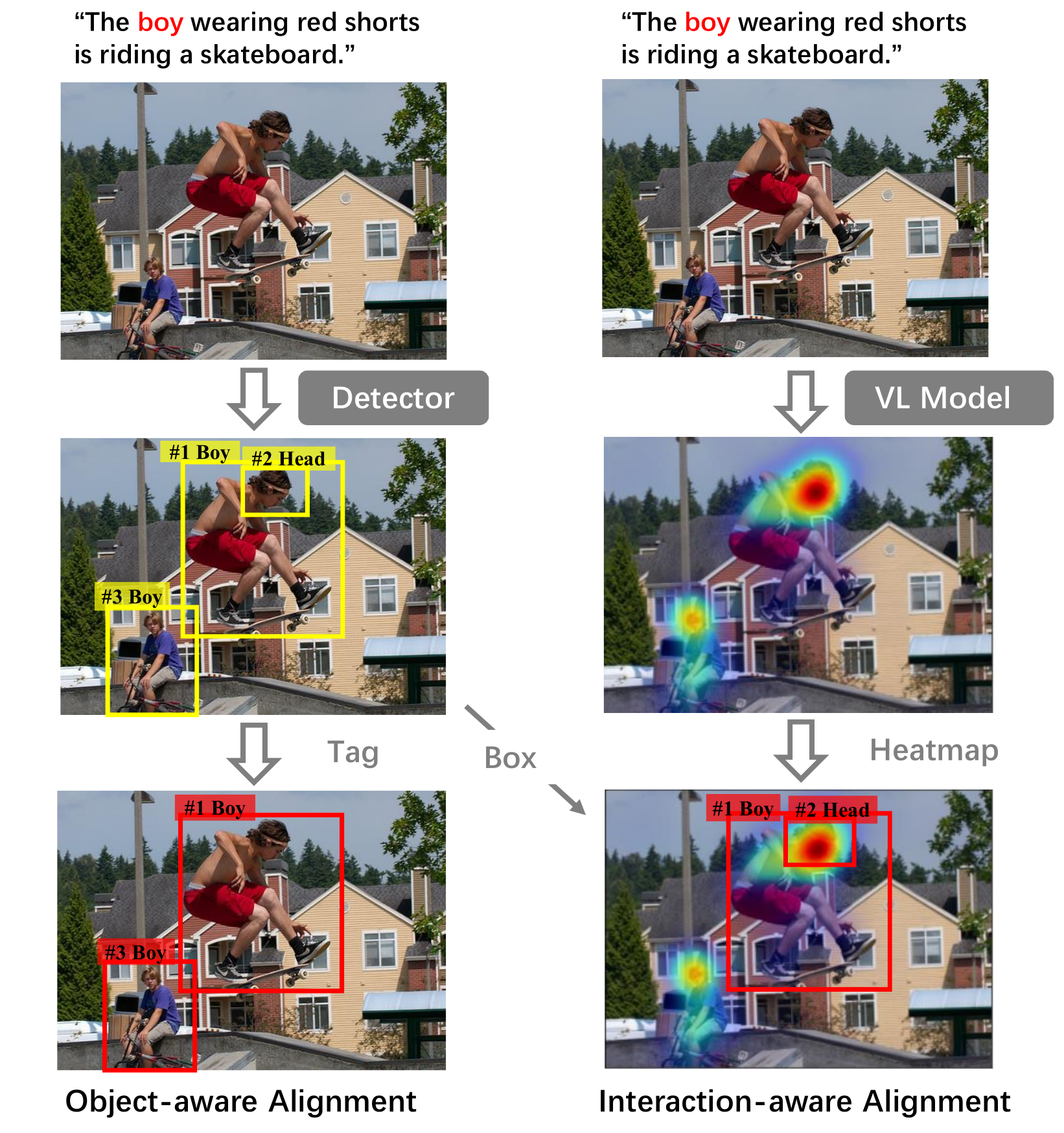}
    \caption{Illustration of the entity-region alignment from object-aware and interaction-aware knowledge.}
    \label{fig:albef}
    \Description[Illustration of the entity-region alignment.]{Illustration of the entity-region alignment.}
\end{figure}

To improve the quality of ``pseudo'' label assignment, one recent WSSGG work~\cite{zhong2021learning} introduces an external knowledge source, \ie the predicted object labels provided by an off-the-shelf object detector. 
A heuristic rule-based algorithm is used to ground text entities to image regions by simply matching entity words with predicted category labels (\eg In Figure~\ref{fig:albef}, Box\#1 and Box\#3 are matched with the entity \texttt{boy}).
Most surprisingly, this simple strategy achieves a new state-of-the-art performance on the WSSGG task.
However, we argue that grounding text entities to image regions merely by matching object categories will neglect the interaction semantics among object pairs. More specifically, the above category-based grounding method lacks the ability to distinguish image regions with the same object category.
For example, there are two boys in the image of Figure~\ref{fig:albef}, and the above mentioned method fails to tell the difference between ``the boy who is riding a skateboard" (\ie Box\#1) and ``the other boy who is sitting" (\ie Box\#3). 

We claim that there are two basic requirements for an ideal alignment between text entities and image regions in WSSGG: 1)~\textbf{Object-consistent}: the grounded regions should have the same object category labels as the entities. 2)~\textbf{Interaction-consistent}: the grounded region pairs should have the same interactions (\ie visual relations) as the entity pairs.
The aforementioned object labels predicted by a pretrained detector can be regarded as a kind of \emph{object-aware} knowledge to help the alignment meet the first requirement.
However, it still neglects the interactions among objects hidden in unlocalized relation triplets or captions.
Thus, we propose to enhance the grounding module with both object-aware and interaction-aware knowledge to acquire more reliable grounding results (\ie ``pseudo'' label assignment). Specifically, we can transfer interaction-aware knowledge from a Vision-Language (VL) model pretrained on large-scale image-caption pairs, which is able to align text tokens to related image areas based on relational reasoning in both textual and visual modality. 
Inspired by~\cite{gao2021towards}, for each text token in a caption, we can construct an activation map that quantitatively measures the relevance between this text token and image locations.
We further utilize this soft alignment between text tokens and image locations as an interaction-aware knowledge source.

Nevertheless, due to the domain gap between these two knowledge sources and SGG datasets, the transferred knowledge is somewhat noisy.
For object-aware knowledge, the off-the-shelf object detector will overlook the interaction semantics in the caption and treat the unseen object categories as background by mistake. For the interaction-aware knowledge, the regions selected by activation maps may actually have inconsistent categories with the text entities. For example in Figure~\ref{fig:albef},
Box\#2 selected by activation map is actually \texttt{head} but matched with \texttt{boy} by mistake.
To alleviate the negative impact of the noise within each knowledge source, we propose to integrate both object-aware and interaction-aware knowledge into the grounding module.
We first design a student grounder with Multiple Instance Learning (MIL) mechanism to learn about in-domain knowledge following the existing weakly supervised visual grounding works~\cite{gupta2020contrastive}.
Meanwhile, to better leverage these two different external knowledge sources (\emph{object-aware} and \emph{interaction-aware}), we regard them as two teachers and fuse their generated targets to guide the training process of the student grounder.
Furthermore, we design two fusing strategies to assign adaptive weights to different teachers for each training sample by assessing the reliability of the generated targets. 

To verify the effectiveness of our proposed method, we conduct extensive experiments and ablation studies on Visual Genome (VG) dataset. Without bells and whistles, our method has greatly improved the performance of WSSGG and achieved new SOTA on various kinds of weak supervision and evaluation metrics.

In summary, we make the following contributions in this paper:
\begin{enumerate}
    \item We propose to enhance the prevalent grounding module in WSSGG with both object-aware knowledge transferred from off-the-shelf object detectors and interaction-aware knowledge transferred from pretrained VL models, which greatly improves WSSGG performance.
    \item We design a knowledge distillation framework to transfer two types of external knowledge to our grounding module to help acquire high-quality ``pseudo'' ground truth for the subsequent SGG training process.
    \item We design two strategies to adaptively assign weights to different external knowledge sources according to their estimated reliability on each training sample.
\end{enumerate}

\section{Related Work}

\paragraph{\textbf{\emph{Weakly supervised SGG}}}
Most current SGG approaches learn to generate scene graphs in a fully-supervised manner~\cite{xu2017scene,chen2019counterfactual,li2017scene,chen2019knowledge,yang2018graph,tian2020part,wang2020memory,tian2021mask,li2022devil,liu2021toward,liu2020adaptively} based on Visual Genome dataset~\cite{krishna2017visual}
As the fully-supervised SGG requires expensive annotations in the form of large quantities of bounding boxes and the enormous number of pairwise relationships, weakly supervised scene graph generation (WSSGG)~\cite{peyre2017weakly,zareian2020weakly,zhang2017ppr} has attracted increasing attention. In~\cite{zhang2017ppr}, researchers firstly proposed to exploit image-level annotations of relation triplets instead of instance-level annotations for SGG learning to reduce the cost of bounding box annotations.
VSPNet~\cite{zareian2020weakly} parsed an image into a semantic
graph by representing objects and predicates as two types of nodes and treating \texttt{subject} and \texttt{object} as two kinds of semantic edges between two types of nodes. Then, they train WSSGG model with a graph alignment algorithm.
Subsequently, some works~\cite{ye2021linguistic,zhong2021learning,shi2021simple} proposed to utilize captions to supervise SGG Learning, which is more natural and easy to be collected.
To convert a caption into SGG supervision, recent works~\cite{ye2021linguistic,zhong2021learning} usually exploit a pretrained language parser~\cite{schuster2015generating} to parse a caption into a set of relationship triplets.
The key of WSSGG is the ideal alignment between text entities and image regions. 
Current WSSGG methods usually train a grounding module under the weak supervision
of aligned image-triplets pairs and received poor performance on the subsequent SGG task.
While one recent work~\cite{zhong2021learning} achieved great improvement by simply utilizing predicted object labels provided by a pretrained object detector to match regions with entities.
In this paper, we integrate two knowledge sources to train a powerful grounding module and generate good-quality ``pseudo" labels for downstream SGG training.

\paragraph{\textbf{\emph{Knowledge Distillation}}}
Knowledge Distillation (KD) aims to transfer information learned from one model to another and has been actively studied in recent years.
It is often characterized by the so-called ``Student-Teacher" (S-T) learning framework.
To make the student learn knowledge from multiple perspectives, recent works focus on distilling knowledge from multiple teachers instead of only one teacher.
One popular method is to KD from the ensemble logits of different teachers~\cite{furlanello2018born, you2017learning, yang2020model, dvornik2019diversity}. In such a way, the student is usually encouraged to learn the averaged outputs of teachers.
However, some works~\cite{xiang2020learning, zhu2018knowledge,chen2022rethinking} argue that purely taking the average of individual outputs may ignore the importance variety of the member teacher, they propose a gating component to control the aggregation of the teacher predictions.
Another prevailing method is to KD from the ensemble of feature representations, which is more flexible and advantageous as it can provide more affluent information.
However, it is more challenging since the feature representation from different teachers is different.
To solve this, one work~\cite{park2019feed} fed the final feature map of the student network into several non-linear layers in parallel and the outputs are trained to mimic the final feature map of the teacher network.
In this paper, we propose a knowledge distillation framework to integrate two external knowledge from pretrained detector and VL model into our grounding module for WSSGG. In addition, we also design two reweight strategies to assign adaptive weights for different knowledge to aggregate them more reasonably.

\begin{figure*}[!ht]
    \centering
    \includegraphics[width=0.95\linewidth]{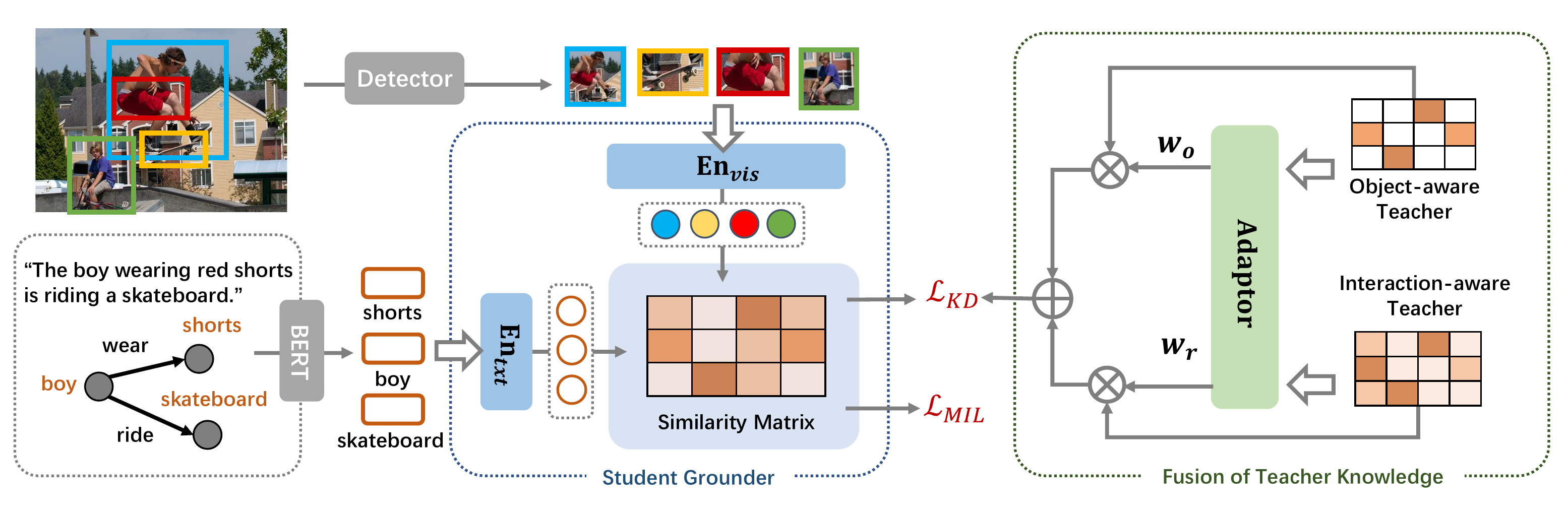}
    \caption{The proposed KD framework to transfer external knowledge to the grounder. \emph{Left}: The architecture of the student grounder. \emph{Right}: The integration of two types of external knowledge. The adaptor represents our fusing strategy to assign adaptive weights to different teachers. We proposed two fusing strategy: 1)~expert-guided reweight and 2)~self-guided reweight.}
    \label{fig:framework}
    \Description[The framework of our proposed model.]{The framework of our proposed model.}
\end{figure*}

\section{Method}
In this section, we introduce the details of our proposed framework for WSSGG. We start with a brief introduction of the pervasive two-staged WSSGG pipeline in Sec.~\ref{sec:3.1}. Then we display the architecture of a simple grounding baseline with MIL loss in Sec.~\ref{sec:3.2}. Lastly, we elaborate on how to integrate object-aware and interaction-aware knowledge into the grounding module via knowledge distillation and adaptive fusion strategies in Sec.~\ref{sec:3.3}.

\subsection{Problem Formulation} \label{sec:3.1}
Given an image $I$, SGG aims to generate a scene graph $\mathcal{G}^{v}=(\mathcal{N},\mathcal{R}) $, where each node in $\mathcal{N}$ corresponds to a bounding box $b \in \mathcal{B}$ with a category $c \in \mathcal{C}$, and each directed edge $r_{so} \in \mathcal{R}$ denotes a predicate for a pair of nodes, pointing from subject $s$ to object $o$. WSSGG differs from fully-supervised SGG in that no bounding box annotation is provided for the nodes in scene graph. Mainstream WSSGG methods typically utilize two types of image-level supervision: captions or unlocalized scene graphs. Caption-based WSSGG methods commonly use a pretrained language parser~\cite{schuster2015generating} to parse the caption into an unlocalized scene graph $\mathcal{G}^{e}=(\mathcal{T},\mathcal{E})$, where each node $t \in \mathcal{T}$ denotes a parsed text entity in captions, and each edge $e_{so} \in \mathcal{E}$ denotes a parsed predicate between entity $s$ and entity $o$. Hereafter, WSSGG methods with both types of supervision can be decoupled into two stages. At the first stage, a grounding module is used to ground each text entity $t \in \mathcal{T}$ in unlocalized graphs to one detected object region $b \in \mathcal{B}$ to generate ``pseudo'' SGG ground truth. At the second stage, a normal SGG model is trained with the generated ``pseudo'' ground truth in a fully-supervised manner.

\subsection{Grounding Baseline with MIL Loss} \label{sec:3.2}
At the grounding stage, following most WSSGG works~\cite{ye2021linguistic,shi2021simple}, we design a simple grounding baseline to encode textual semantic features and image visual features into the same space and calculate cross-modal similarities to acquire the grounding results.

\subsubsection{\textbf{Visual and Textual Representations}}
Due to the domain gap between visual and textual features, we use two separate encoders to map them into the same feature space.

\textbf{Text Encoder.}
Given a caption, we exploit a language parser~\cite{schuster2015generating} to extract a set of text entities from the caption, $\mathcal{T}=\{t_i\}_{i=1}^{N_e}$, where $N_e$ is the number of entities. To properly represent text entities and better preserve the contextual information from the caption, we feed the entire caption into a pretrained BERT model~\cite{devlin2019bert} and utilize the final hidden states at corresponding places as contextualized embeddings for extracted entities, denoted as $\Mat{E} \in \mathbb{R}^{N_e \times d_e}$. After that, we send the contextualized embeddings of entities into a text encoder for cross-modal alignment:
\begin{equation}\label{eq:Htxt}
    \Mat{H} = \text{En}_{txt}(\Mat{E}),
\end{equation}
where $\text{En}_{txt}$ is a two layer MLP, $\Mat{H} = [\Mat{h}_1, \Mat{h}_2, \cdots, \Mat{h}_{N_e}]^T$ is the encoded embeddings of text entities, and $\Mat{h} \in \mathbb{R}^d$.

\textbf{Visual Encoder.}
For each image, we adopt a pretrained object detector to generate a set of region proposals $\mathcal{B}=\{b_i\}_{i=1}^{N_v}$ with corresponding visual features $\Mat{X} \in \mathbb{R}^{N_v \times d_v}$, where $N_v$ is the number of proposals.
To map the proposal visual features into the same space as text features, we feed them into a visual encoder $\text{En}_{vis}$:
\begin{equation} \label{eq:Hvis}
    \Mat{V} = \text{En}_{vis}(\Mat{X}),
\end{equation}
where $\text{En}_{vis}$ is a two layer MLP, $\Mat{V} = [\Mat{v}_1, \Mat{v}_2, \cdots, \Mat{v}_{N_v}]^T$ is the encoded embeddings of region proposals, and $\Mat{v} \in \mathbb{R}^{d}$.

\subsubsection{\textbf{Visual-Textual Similarity}}
For each image-caption pair, after obtaining the textual embeddings $\Mat{H}$ for text entities and visual embedding $\Mat{V}$ for region proposals, we construct an similarity matrix $\Mat{A}=[\Mat{a}_1, \Mat{a}_2, ..., \Mat{a}_{N_e}]^T$, where $\Mat{a}_i \in \mathbb{R}^{N_v}$ represents the similarity vector between entity $t_i$ and all region proposals:
\begin{equation} \label{eq:attentionmat}
    \Mat{a}_{i} = [a_{i,1}, a_{i,2}, \cdots, a_{i,N_v}]^T, \quad a_{i,j}  = \text{FC}_{att}(\Mat{h}_i \odot \Mat{v}_j),
\end{equation}
where $\odot$ denotes element-wise production, and $\text{FC}_{att}$ is a FC layer to calculate the similarity scores between entities and proposals.

\subsubsection{\textbf{Training with MIL}}
We first exploit Multiple Instance Learning (MIL) mechanism to learn about in-domain knowledge from image-caption supervision.
The image can be viewed as a bag of region proposals, and the caption can be viewed as a bag of entities. For each caption, the paired image is regarded as the positive bag as it is guaranteed to contain region proposals that are matched with the given caption while all other images are regarded as negative bags. During training, for each caption, we take the paired image as the positive sample, all other images in the mini-batch as negative samples, and adopt a triplet loss to train the grounding baseline.

Specifically, for each image-caption pair $(I, C)$, we compute the similarity matrix $\Mat{A} \in \mathbb{R}^{N_e \times N_v}$ between all parsed entities and region proposals. Afterwards, for each entity $t_i$, we take the maximum value of $\Mat{a}_{i}$ to represent the similarity between entity $t_i$ and image $I$, denoted as $S(t_i, I)$. Then we take the average over all entities to obtain the overall similarity score between image $I$ and caption $C$:
\begin{equation}
    S(C, I) = \frac{1}{N_e} \sum_{i=1}^{N_e} S(t_i, I), \quad S(t_i, I) = \max (\sigma(\Mat{a}_i)),
\end{equation}
where $\sigma$ is sigmoid function to map values to $(0, 1)$.

Finally, we utilize a triplet loss to train the grounding module:
\begin{equation}
\mathcal{L}_{MIL} =  \sum_{C, I, I'}\max(0, S(C, I')-S(C, I)+\alpha_1 ),
\end{equation}
where $\alpha_1$ is a predefined margin, $S(C, I)$ is the similarity between caption $C$ and positive image $I$, while $S(C, I')$ is the similarity between caption $C$ and negative image $I'$ within the minibatch.

\subsection{Integrating External Knowledge} \label{sec:3.3}
The grounding module proposed in Sec.~\ref{sec:3.2} is simple yet efficient. However, we argue that an ideal alignment between text entities and image regions should be both object-consistent and interaction-consistent. It's hard to meet these requirements merely with the MIL training scheme. To improve the quality of grounding results, we propose to enhance our grounding module with both object-aware and relation-aware knowledge. Specifically, for each entity $t_i$ we firstly generate targets from two external knowledge, denoted as $\hat{\Mat{a}}_i^{(o)}$ and $\hat{\Mat{a}}
_i^{(r)}$. Then, we design a Multi-teacher Knowledge Distillation framework to fuse these two targets and use the fused targets $\Mat{q}
_i$ to guide the output distribution of the grounding module.

\subsubsection{\textbf{Object-aware Knowledge}}
We believe that the off-the-shelf object detector pretrained on a large-scale dataset is a natural expert at categorical recognition of objects. Hence we exploit the predicted object category of the region proposals as external knowledge to help ensure the object-consistent property of the grounding results. Given a text entity $t_i$, for each region proposal $b_j$ with predicted category $c_j$, if $c_j$ is matched with $t_i$ (\ie the same word or synonyms), the corresponding place in the target vector is set as $1$, otherwise $0$. We perform linear scaling to assure the target vector sums to $1$. Formally, the generated target for entity $t_i$ is:
\begin{equation}
    \hat{\Mat{a}}^{(o)}_{i} = [\hat{a}_{i,1}^{(o)}, \hat{a}_{i,2}^{(o)}, ..., \hat{a}_{i,N_v}^{(o)}]^T, \quad \hat{a}_{i,j}=\frac{ \mathds{1}[t_i=c_j]}{ {\textstyle \sum_{k=1}^{N_v}} \mathds{1}[t_i=c_k] } ,
\end{equation}
where $\mathds{1}[condition]$ is $1$ if $condition$ is true, otherwise 0.

\subsubsection{\textbf{Interaction-aware Knowledge}}
To integrate interaction-aware knowledge, we propose to transfer knowledge from pretrained visual-language (VL) models which have shown promising results on complex visual-language relational reasoning tasks. For a pair of image $I$ and caption $C$, the pretrained VL model~\cite{li2021align} normally contains multiple cross-attention layers to measure the relevance between each image location and each token in the caption. We can visualize the activation maps of such attention layers with Grad-CAM~\cite{selvaraju2017grad}, as shown in Figure~\ref{fig:albef}. Inspired by~\cite{gao2021towards}, we utilize these activation maps as guidance to construct interaction-aware targets.

Specifically, for each entity $t_i$ in the caption, we exploit Grad-CAM to construct an activation map $\Mat{\Phi}_i$, representing the relevance between the text entity and each image location. Then, following~\cite{gao2021towards}, the relevant score of region proposal $b_j$ is calculated as:
\begin{equation}
    g_{i,j} = \frac{\Mat{\Phi}_i (b_j)}{\sqrt{|b_j|} },
\end{equation}
where $\Mat{\Phi}_i (b_j)$ denotes the summation of entity $t_i$'s activation map within region proposal $b_j$, and $|b_j|$ denotes the area of proposal $b_j$. The relevant score $g_{i,j}$ can be interpreted as the the activation density of region proposal $b_j$ with respect to entity $t_i$.

Finally, we normalize relevant scores of all regions with softmax to get the interaction-aware target for entity $t_i$, denoted as $\hat{\Mat{a}}^{(r)}_{i}$:
\begin{equation}
    \hat{\Mat{a}}^{(r)}_{i} = [\hat{a}_{i,1}^{(r)}, \hat{a}_{i,2}^{(r)}, ..., \hat{a}_{i,N_v}^{(r)}]^T, \quad \hat{a}_{i,j}^{(r)} = \frac{\text{exp}(g_{i,j})}{ {\sum_{k=1}^{N_v}}\text{exp}(g_{i,k}) } .
\end{equation}

\subsubsection{\textbf{Fusing Strategy}}
After obtaining generated targets of two teachers, one intuitive fusing strategy is to simply treat each teacher equally and average the generated targets. However, due to the domain gap between these two knowledge and SGG dataset, the generated targets are inevitably noisy. To some extent, this negative influence of noisy labels can be alleviated by mining the complementary information between these two knowledge. Therefore, we propose to assign adaptive weights to each target and then the fused target is calculated by the weighted sum:
\begin{equation} \label{eq:label-fusion}
    \Mat{q}_i = w^{(o)}*\hat{\Mat{a}}^{(o)}_{i} + w^{(r)}*\hat{\Mat{a}}^{(r)}_{i}.
\end{equation}

For each entity $t_i$, we aggregate the encoded visual features of all region proposals weighted by the generated target $\hat{\Mat{a}}^{(o/r)}_i$, which can be regarded as the attended visual features of entity $t_i$ in the view of object-aware or interaction-aware teacher, denoted as $\widetilde{\Mat{v}_i}^{(o/r)}$:
\begin{equation} 
    \widetilde{\Mat{v}_i}^{(o/r)} = \Mat{V}^T \cdot \hat{\Mat{a}}_i^{(o/r)},
\end{equation}
where $\Mat{V}$ is the encoded proposal visual features according to Eq.~\eqref{eq:Hvis}.

Then, we compute the similarities between the attended visual features and textual feature of entity $t_i$ to estimate the reliability of each teacher with respect to entity $t_i$:
\begin{equation}\label{eq:sim}
    sim_i^{(o/r)} = f(\widetilde{\Mat{v}_i}^{(o/r)}, \Mat{h}_i),
\end{equation}
where $\Mat{h}_i$ is the textual feature for entity $t_i$ according to Eq.~\eqref{eq:Htxt}.

After normalizing, the reliability scores of the teachers are used as the adaptive weights in Eq.~\eqref{eq:label-fusion}. As for the similarity metric, we propose two alternatives: expert-guided pre-computed weights from an off-the-shelf image-text matching model or self-guided learnable weights trained with triplet loss.

\paragraph{\emph{\textbf{Expert-guided Reweight}}} To estimate the similarity between attended visual feature and entity textual feature, one easy way is to utilize an off-the-shelf image-text matching model. We experimented with the CLIP model~\cite{radford2021learning} which is pretrained on large-scale image-text datasets and has proven to be a powerful tool in many visual-language tasks. We first fit the text entity into a prompt sentence and use CLIP to extract textual features. Then we use CLIP to extract visual features of the cropped image regions and construct the attended visual feature for each teacher. Finally, we compute the cosine similarity between the encoded features. Note that $\Mat{V}$ in Eq.~\eqref{eq:label-fusion} and $\Mat{h}_i$ in Eq.~\eqref{eq:sim} are replaced by CLIP encoded features.

\begin{table*}
    \centering
    \caption{SGDet Results on three types of weak supervison: Unlocalized Graph, VG Caption and COCO Caption. Motifs~\cite{zellers2018neural} and Uniter~\cite{chen2020uniter} are two fully-supervised SGG models. \textsuperscript{$\dagger$} denotes our re-implemented results using the official released code.}
    \scalebox{0.95}
    {
    \begin{tabular}{ c | c | c |c | l | l l l  }
    \hline
     \multicolumn{3}{c|}{Training Setting}  & \multirow{2}*{ SGG model}& \multirow{2}*{Methods} & \multicolumn{3}{c}{ SGDet} \\
        Supervision &  $\#$ Triplets & $\#$ Images &  &  & R@20 & R@50 & R@100 \\ 
     \hline
     \multirow{7}*{\tabincell{c}{Unlocalized\\Graph}}  & \multirow{7}*{406K} &  \multirow{7}*{58K} &  ---&VSPNet~\cite{zareian2020weakly} & --- &4.70 & 5.40  \\
      &  &     &  ---  & LSWS~\cite{ye2021linguistic}  & --- &7.30 & 8.73 \\
      &   &    & Motifs & WSGM~\cite{shi2021simple} & 4.12 &5.59 & 6.45 \\ 
      &   &    & Motifs & SGNLS~\cite{zhong2021learning}\textsuperscript{$\dagger$} & 7.23 & 9.28 & 10.71 \\
      &   &    & Motifs  & \textbf{Ours} &$\textbf{9.09}_{\textcolor{red}{+1.86}}$ &  $\textbf{11.39}_{\textcolor{red}{+2.11}}$ &  $\textbf{12.89}_{\textcolor{red}{+2.18}}$ \\ 
      \cdashline{4-8}[1pt/1pt]
      &   &   &   Uniter & SGNLS~\cite{zhong2021learning}\textsuperscript{$\dagger$} & 7.81 & 10.03 & 11.50 \\ 
     &   &   &   Uniter & \textbf{Ours} &   $\textbf{9.57}_{\textcolor{red}{+1.76}}$ & $\textbf{11.80}_{\textcolor{red}{+1.77}}$ &  $\textbf{13.15}_{\textcolor{red}{+1.65}}$ \\ 
     \hline
      \multirow{5}*{VG Caption}  & \multirow{5}*{498K}  & \multirow{5}*{62K} & ---  & LSWS~\cite{ye2021linguistic} & --- & 3.85 & 4.04 \\ 
      &  &   & Motifs &   SGNLS~\cite{zhong2021learning}\textsuperscript{$\dagger$} & 6.31 & 8.05 & 9.21   \\
    &   &   & Motifs & \textbf{Ours} & $\textbf{8.25}_{\textcolor{red}{+1.94}}$ &  $\textbf{10.50}_{\textcolor{red}{+2.45}}$ &  $\textbf{11.98}_{\textcolor{red}{+2.77}}$ \\ 
      \cdashline{4-8}[1pt/1pt]
      &  &     & Uniter & SGNLS~\cite{zhong2021learning}  &--- & 9.20 & 10.30  \\ 
    &  &   &  Uniter   &  \textbf{Ours}  &  \textbf{8.90} &  $\textbf{10.93}_{\textcolor{red}{+1.73}}$ &  $\textbf{12.14}_{\textcolor{red}{+1.84}}$ \\ 
     \hline
     \multirow{4}*{COCO Caption} & \multirow{4}*{131K} & \multirow{4}*{57K} &--- & LSWS~\cite{ye2021linguistic} &--- & 3.28 &  3.69 \\ 
            &   &    & Motifs & \textbf{Ours} & 5.02 & 6.40 & 7.33  \\
    \cdashline{4-8}[1pt/1pt]
      &   &   & Uniter & SGNLS~\cite{zhong2021learning} & --- &  5.80  & 6.70 \\ 
      &   &    &   Uniter & \textbf{Ours} & \textbf{5.42} & $\textbf{6.74}_{\textcolor{red}{+0.94}}$  & $\textbf{7.62}_{\textcolor{red}{+0.92}}$ \\ 
    \hline
    \end{tabular}
    }
    \label{tab:comparision}
\end{table*}

\paragraph{\emph{\textbf{Self-guided Reweight}}}  
We also design another learnable adaptor network to assign weights to different teachers. For each entity $t_i$, we estimate the reliability of the generated target by:
\begin{equation}
    sim^{(o/r)}_i = \text{FC}_{adp}(\widetilde{\Mat{v}}^{(o/r)}, \Mat{h}_i).
\end{equation}
We note that the gradient will not pass through $ \widetilde{\Mat{v}}^{(*)}$ in order not to affect the visual feature encoder in the grounding module.

To train the adaptor network, for each entity, we regard the attend visual features of either teacher as positive samples and randomly attended visual features as negative samples. We expect the adaptor network to assign higher weights to positive samples as compared to negative samples. To accomplish this, we use the triplet loss to train the adaptor network:
\begin{equation}
    \mathcal{L}_{adp} = \sum_{i}\max(0, sim^{(neg)}_i-sim^{(o/r)}_i+\alpha_2 ),
\end{equation}
where $\alpha_2$ is a predefined margin, $sim^{(neg)}_i$ is the reliability score between entity $t_i$ with some randomly attended visual feature.

\subsubsection{\textbf{Training with Knowledge Distillation}}
After fusing the generated targets from different teachers, we expect the output similarity scores of our grounding module to be close to the fused targets. We adopt KL divergence loss for knowledge distillation:
\begin{equation}
    \mathcal{L}_{KD} = \sum^{N_e}_{i=1} \text{KL}(\Mat{q}_i||\Mat{p}_i), \quad \Mat{p}_i = \text{Softmax}(\Mat{a}_i),
\end{equation}
where $\Mat{a}_i$ is the similarity vector of entity $t_i$ output by grounding module (\cf Eq.~\eqref{eq:attentionmat}), $\Mat{q_i}$ is the fused targets from teachers for $t_i$.

The total training loss of grounding module is:
\begin{equation}
    \mathcal{L}_{GD} = \mathcal{L}_{MIL} + \mathcal{L}_{KD} + \lambda \mathcal{L}_{adp},
\end{equation}
where $\lambda$ is set to $1$ when the learnable weights are used and $0$ when pre-computed weights are used.

\subsection{SGG Training with Pseudo Ground Truth}
After training, our grounding module can be used to compute similarity scores between image region proposals and entities in captions or unlocalized scene graphs. For each entity in the training set, we rank all region proposals of the same image and select top-K (\eg 3) regions with the highest similarity scores as region candidates. These grounding results together with the unlocalized scene graphs constitute ``pseudo'' ground truth which can be used to train an arbitrary SGG model in a fully supervised manner:
\begin{equation}
    \mathcal{L}_{SGG} = \mathcal{L}_{obj} + \mathcal{L}_{rel},
\end{equation}
where $\mathcal{L}_{obj}$ and $\mathcal{L}_{rel}$ are both cross entropy loss for object and relation classification, respectively. At each training step, we randomly select one candidate to promote robustness.

\section{Experiment}
In this section, we conduct extensive experiments to verify the effectiveness of our proposed method for WSSGG.

\subsection{Experimental Settings}
\subsubsection{\textbf{Datasets}} 
We conduct our experiments on two benchmarks: \textbf{Visual Genome} (VG)~\cite{krishna2017visual} and \textbf{COCO}~\cite{chen2015microsoft}:
1) VG is a large-scale dataset commonly used in the SGG task. Following most previous works~\cite{zellers2018neural, yan2020pcpl,zhong2021learning,zareian2020weakly}, we adopt popular VG split~\cite{zellers2018neural, zhong2021learning}, which includes most frequent 150 object categories and 50 predicate classes.
After preprocessing, each image has $11.5$ objects and $6.2$ relationships on average. The split uses $70\%$ of images for training (including $5K$ images for validation) and the rest $30\%$ for test.
2) COCO~\cite{chen2015microsoft} contains 123K images. Each image corresponds to 5 human-annotated captions. Following~\cite{zhong2021learning}, we selected 106k images
in COCO for training by filtering out images that exist in the
test set of VG.
All the experiments are evaluated on VG test split.

\subsubsection{\textbf{Tasks}}
To compare WSSGG performance, we followed prior WSSGG works~\cite{ye2021linguistic,zhong2021learning,zareian2020weakly} and mainly evaluated the performance on \textbf{Scene Graph Detection (SGDet)} task: Given an image, models are required to detect object classes and localization, and predict predicate classes for an object pair.
Meanwhile, we followed~\cite{ye2021linguistic,zhong2021learning} and compared WSSGG performance on three types of weak supervision settings:
1)~\textbf{Unlocalized Graph}: Models are trained on image-level relation triplet annotations without object bounding box annotations.
2)~\textbf{VG Caption}: Models are trained on image-caption pairs provided in VG training split.
3)~\textbf{COCO Caption}: Models are trained on image-caption pairs in COCO dataset with the images in VG test split removed.

\subsubsection{\textbf{Evaluation Metrics}}
To compare with other SGG methods, we mainly reported the performance on Recall@K, which measures the fraction of ground truth relationship triplets that appear among the top $K$ most confident triplet predictions in an image. 
For a more complete comparison, we also reported object detection mAP to compare the object prediction performance in the ablation study.

\subsubsection{\textbf{Implementation Details}}
Following~\cite{zhong2021learning}, we used a Faster R-CNN detector~\cite{ren2015faster} pretrained on OpenImages~\cite{alina2020open}, capable of detecting 601 object categories. We kept the top 36 objects per image and extracted the 1536-D region features from the detector.
We also followed~\cite{ye2021linguistic,zhong2021learning} and exploited a language parser in~\cite{schuster2015generating} to convert captions paired with images into a set of relation triplets.
For interaction-aware knowledge transfer, we exploited a visual-language model ALBEF~\cite{li2021align}, which is pretrained on a large collection of image-caption pairs without bounding box annotations~\cite{li2021align,gao2021towards}.
In practice, we set the value of both $\alpha_1$ and $\alpha_2$ as $0.2$, 
and we set the learning rate as $0.001$ for all models on unlocalized graph setting and as $0.0001$ on two caption-supervised settings.

\subsection{Comparison with State-of-the-Arts}
We report our main results on SGDet and compare performance with other WSSGG models to show the effectiveness of our model.

\begin{itemize}
    \item \textbf{VSPNet}~\cite{zareian2020weakly} represented objects and predicates as two types of nodes and treated \texttt{subject} and \texttt{object} as two kinds of semantic edges between them. Then they trained the weakly supervised SGG model with a graph alignment algorithm.
    \item \textbf{LSWS}~\cite{ye2021linguistic} designed an attention matrix to complete the alignment between text graph and detected objects. Then they send the matching labels to guide SGG training process and improved the alignment through iterative refinement.
    \item \textbf{SGNLS}~\cite{zhong2021learning} directly utilized the detection tags provided by detector to generate pseudo labels. The generated pseudo labels were sent to a fully-supervised SGG model.
    \item \textbf{WSGM}~\cite{shi2021simple} proposed a graph matching algorithm and conduct a contrastive learning framework based on matching results to train a grounding module. Then the grounding results were sent to a fully-supervised model.
\end{itemize}

\textbf{Results on Unlocalized Graph Setting.}
From the top of Table~\ref{tab:comparision}, we can see that our method set a new record on Unlocalized Graph Setting and exceeded the strongest baseline SGNLS with significant improvement.
Particularly, our method has achieved $22.5\%$ (9.57 \vs 7.81) improvement relatively over the strongest baseline SGNLS on R@20 equipped with Uniter, and achieved $25.7\%$ (9.09 \vs7.23) improvement over SGNLS on R@20 with Motifs.

\textbf{Results on VG Caption Setting.}
Caption supervision is weaker and more challenging than unlocalized Graph setting. 
From the middle of Table~\ref{tab:comparision}, we can observe that our method still achieved the best performance.
Especially, Our method outperforms SGNLS by $30.4\%$ ($10.50$ \vs $8.05$) relatively on R@50 equipped with Motifs and outperforms SGNLS by  $18.8\%$ (10.93 \vs 9.20) on R@50 equipped with Uniter.
This is owing to that our integrated interaction-aware knowledge takes the best advantage of interaction semantics hidden in captions, resulting in high-quality grounding results.

\textbf{Results on COCO Caption Setting.}
COCO Caption is the most difficult setting in weak supervision due to the distribution shift existing in different datasets.
Even though, our method still achieved outstanding performance as shown at the bottom of Table~\ref{tab:comparision}.
Compared with SGNLS, our method with Uniter is superior to SGNLS by $16.2\%$ (6.74 \vs 5.80) improvement on R@50.

\subsection{Ablation Study}
We validate the effectiveness of our model by answering the following questions: \textbf{Q1}: How do both types of integrated knowledge contribute to the performance improvement? \textbf{Q2}: How do the two proposed fusing strategies perform? \textbf{Q3}: Does our model provide more reliable ``pseudo'' ground truths for SGG training?

\subsubsection{\textbf{External Knowledge Sources (Q1)}}
To verify the effectiveness of each knowledge source we integrated, we report SGDet performance of Motifs~\cite{zellers2018neural} model on VG Caption supervision when integrating different knowledge sources, which is shown in Table~\ref{tab:teacher}. 
MIL loss is designed to learn about in-domain knowledge of weak supervision data.
Compared the results in the 1st line of Table~\ref{tab:teacher} and the 8th line of Table~\ref{tab:comparision}, we can see that grounding module with our simple MIL Loss (MIL) can achieve similar performance with elaborately designed model LSWS~\cite{ye2021linguistic}. We then progressively integrate object-aware knowledge (Object-K) and interaction-aware knowledge (Interact-K) to the grounding module.
From Table~\ref{tab:teacher}, we can observe that integrating either Object-aware Knowledge (2nd line) or interaction-aware Knowledge (3rd line) is able to enhance model performance with great improvement compared to baseline (1st line), which verifies the effectiveness of two knowledge. And integrating two knowledge together (4th line) achieved the best performance. This is owing to that the two knowledge teachers are complementary to each other, and our grounding module is capable of taking full advantage when learning from both of them together.

\begin{table}
    \centering
    \caption{SGDet (VG-caption) results of Motifs~\cite{zellers2018neural} by integrating different knowledge into the baseline grounding module trained with MIL loss (MIL): Object-aware Knowledge (Object-K), interaction-aware Knowledge (Interact-K).}
    \scalebox{0.95}
    {
    \begin{tabular}{ c c  c | c c c}
    \hline
    \multirow{2}*{MIL} & \multirow{2}*{Object-K} & \multirow{2}*{Interact-K} & \multicolumn{3}{c}{SGDet} \\
     & & & R@20& R@50 & R@100 \\
    \hline
     \ding{51} &             &           & 2.79 & 3.63  & 4.37 \\
    \ding{51} & \ding{51} &             & 6.42 & 8.41 & 9.86   \\
    \ding{51} &             &\ding{51} & 6.40 & 8.52 & 9.98   \\
    \ding{51} &\ding{51} &\ding{51} & \textbf{8.14} & \textbf{10.23} & \textbf{11.48}  \\
    \hline
    \end{tabular}
    }
    \label{tab:teacher}
\end{table}

\begin{table}
    \centering
    \caption{SGDet (VG-caption) results of Motifs~\cite{zellers2018neural} based on our grounding module with different fusing strategies for the generated targets from different teachers.}
    \scalebox{0.95}
    {
    \begin{tabular}{ l | c c  c}
    \hline
    \multirow{2}*{Fusing Strategy} &  \multicolumn{3}{c}{SGDet} \\
    & R@20& R@50 & R@100 \\
    \hline
    Average & 8.14  & 10.23  & 11.48 \\           
    Self-guided Reweight & \textbf{8.27}  & 10.38  & 11.78  \\
    Expert-guided Reweight & 8.25  & \textbf{10.50}  & \textbf{11.98}  \\
    \hline
    \end{tabular}
    }
    \label{tab:fusing}
\end{table}

\begin{table}
    \centering
    \caption{Object Detection mAP Performance Comparison. }
    \scalebox{0.95}
    {
    \begin{tabular}{ c | c | l |c  }
    \hline
    Supervision & SGG Model & Methods & \tabincell{c}{Object Detection\\mAP} \\
    \hline
    \multirow{4}*{ \tabincell{c}{Unlocalized\\Graph}} &  Uniter & SGNLS & 10.74\\
     &  Uniter & \textbf{Ours} & \textbf{14.46} \\
    \cdashline{2-4}[1pt/1pt]
     &  Motifs & SGNLS\textsuperscript{$\dagger$}  &11.72\\
     & Motifs & \textbf{Ours} &\textbf{14.90} \\
    \hline
     \multirow{3}*{VG Caption}  & Uniter & \textbf{Ours} &14.12 \\
     \cdashline{2-4}[1pt/1pt]
     & Motifs &   SGNLS          & 10.22\\
     & Motifs &   \textbf{Ours}  & \textbf{14.50} \\
    \hline
    \end{tabular}
    }
    \label{tab:detection}
\end{table}

\begin{figure*}
    \centering
    \includegraphics[width=0.92\linewidth]{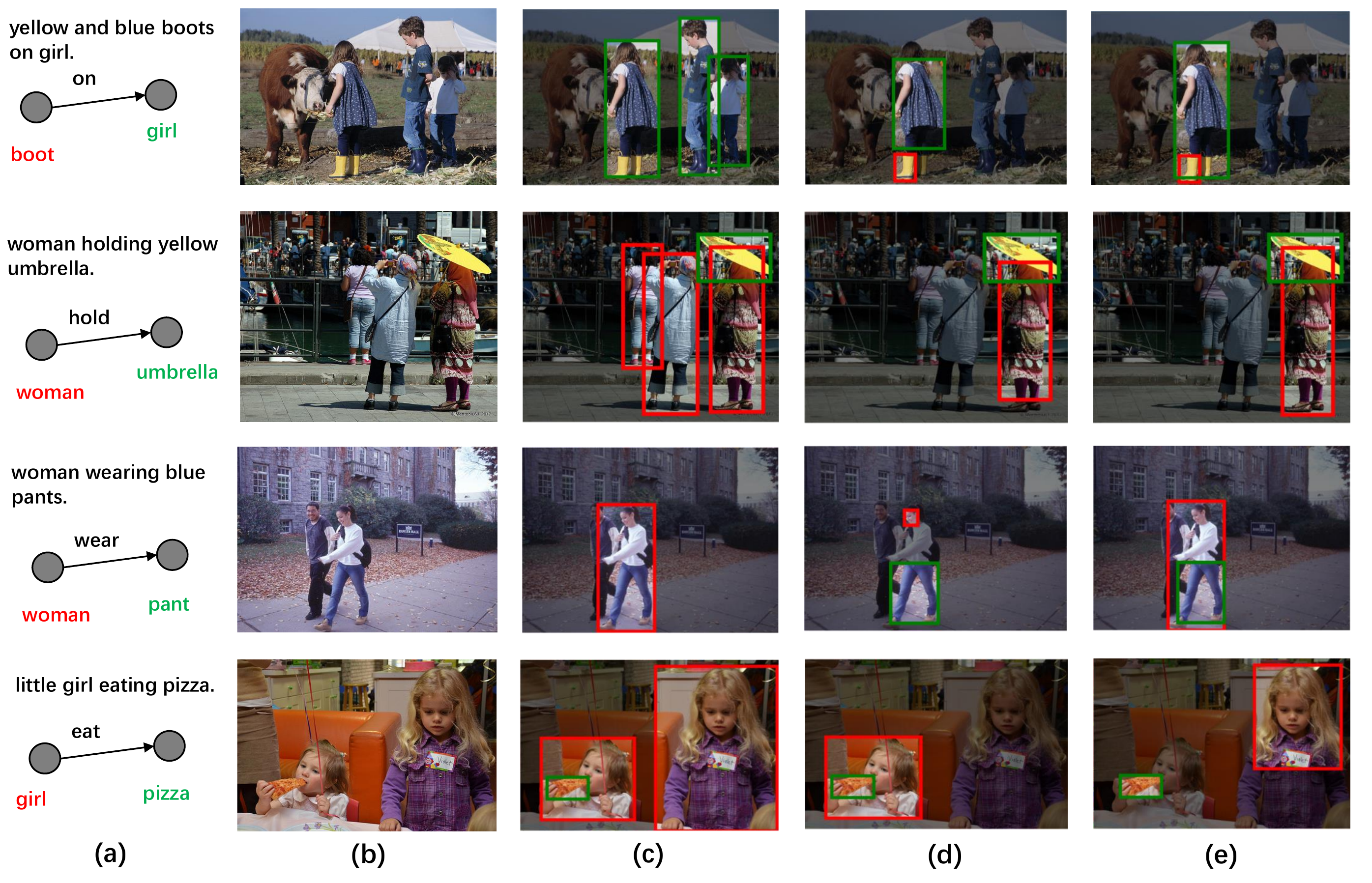}
    \caption{Examples of grounding results output by our model and knowledge teachers. (a):~The caption and parsed triplets. (b):~Input image. (c):~Grounding results of object-aware teacher. (d):~Top-1 grounding results of interaction-aware teacher. (e):~Top-1 grounding results of our model. The grounded subjects and objects are drawn by \textcolor{ForestGreen}{green} and \textcolor{red}{red} boxes respectively. }
    \label{fig:visualization}
\end{figure*}

\subsubsection{\textbf{Target Fusing Strategy (Q2)}}
We design two fusing strategies for aggregating targets from different knowledge teachers. To compare them, we report the performance of Motifs~\cite{zellers2018neural} based on our grounding module with different fusing strategies in Table~\ref{tab:fusing}. 
From Table~\ref{tab:fusing}, we can observe that both self-guided reweight and expert-guided reweight can improve the model performance. We believe the improvement is due to the fact that adaptively assigning weights to different targets according to their reliability is able to alleviate the negative effect of the internal noise within different knowledge sources and help acquire high-quality grounding results.
Compared to expert-guided strategy, self-guided strategy does not require external guidance but still improved performance compared to the simple averaging strategy.

\subsubsection{\textbf{Detection Performance (Q3)}}
We report Object Detection mAP performance in Table~\ref{tab:detection} and compare with a strong baseline SGNLS~\cite{zhong2021learning}, which utilized the knowledge from a well-pretrained detector
We can observe that SGG models trained with our generated ``pseudo'' labels perform even better on object detection task. 
Our method with Motifs is superior to SGNLS by $27.1\%$ (14.90 \vs 11.72) and $41.8\%$ (14.50 \vs 10.22) on Unlocalized Graph and VG Caption respectively.
That is due to that our method is capable of handling unseen object categories by detector, benefiting from integrating interaction-aware knowledge.

\subsubsection{\textbf{Case Study (Q3)}}
We display some cases output by our grounding module and two knowledge teachers, as shown in Figure~\ref{fig:visualization}. 
For object-aware teacher, we display all grounding outputs, while for interaction-aware teacher and our grounding module, we only display Top-1 grounding results for clear visualization.
We can see that object-aware teacher is able to find the object-consistent regions for the target entity, but it is unable to tell the difference among the regions with different interactions (\cf 1st and 3rd examples). Additionally, it failed to match regions when it comes to unseen object categories (\eg \texttt{boot} has no matched region in 1st example).
While for interaction-aware teacher, it is able to find the relevant region for the target entity, but the output regions may have inconsistent object category with the target entity (\eg \texttt{woman} is grounded to the \texttt{dress} in 1st example, and grounded to the \texttt{head} in 3rd example).
In comparison, our method is more likely to generate grounding results that are both object-consistent and interaction-consistent.
For fairness, we also display a failure case of our method (\cf 4th example).

\section{Conclusion}
In this paper, we integrated both object-aware knowledge and interaction-aware knowledge into our grounding module for Weakly supervised Scene Graph Generation (WSSGG) task to acquire high-quality ``pseudo'' ground truth for the downstream ``fully-supervised" SGG training.
We designed a knowledge distillation framework to leverage these two types of knowledge, which regards them as two teachers and aggregates their generated targets with adaptive weights by assessing their reliability on each training sample.
Extensive experiments on Visual Genome and COCO benchmarks demonstrate the effectiveness of our proposed model.
Moving forward, we plan to assign soft ``pseudo'' labels instead of hard labels as soft labels may provide more affluent information for SGG training.
We can also extend our model to the video domain and apply it to weakly supervised video relation tasks~\cite{gao2021video,gao2022classification}.
\begin{acks}
This work was supported by the National Key Research \& Development Project of China (2021ZD0110700), the National Natural Science Foundation of China (U19B2043, 61976185), Zhejiang Natural Science Foundation (LR19F020002), Zhejiang Innovation Foundation (2019R52002), and the Fundamental Research Funds for the Central Universities (226-2022-00087).
\end{acks}


\bibliographystyle{ACM-Reference-Format}
\bibliography{refer}

\appendix

\end{document}